\documentclass{article}
\usepackage{iclr2026_conference,times}

\usepackage{hyperref}
\usepackage{url}
\usepackage{amsmath,amssymb}
\usepackage{graphicx}
\usepackage{booktabs}
\usepackage{multirow}
\usepackage{xcolor}
\usepackage{cleveref}
\usepackage{microtype}
\usepackage{enumitem}
\usepackage{nicefrac}

\newcommand{\crps}{\text{CRPS}}

\iclrfinalcopy

\track{Research}

\title{Dissecting Chronos: Sparse Autoencoders Reveal Causal Feature Hierarchies in Time Series Foundation Models}

\author{%
Anurag Mishra\\
Rochester Institute of Technology\\
\texttt{am2552@rit.edu}
}

\begin{document}

\maketitle

\begin{abstract}
Time series foundation models (TSFMs) are increasingly deployed in high-stakes domains, yet their internal representations remain opaque. We present the first application of sparse autoencoders (SAEs) to a TSFM, training TopK SAEs on activations of Chronos-T5-Large (710M parameters) across six layers. Through 392 single-feature ablation experiments, we establish that every ablated feature produces a positive CRPS degradation, confirming causal relevance. Our analysis reveals a depth-dependent hierarchy: early encoder layers encode low-level frequency features, the mid-encoder concentrates causally critical change-detection features, and the final encoder compresses a rich but less causally important taxonomy of temporal concepts. The most critical features reside in the mid-encoder (max single-feature $\Delta\crps = 38.61$), not in the semantically richest final encoder layer, where progressive ablation paradoxically improves forecast quality. These findings demonstrate that mechanistic interpretability transfers effectively to TSFMs and that Chronos-T5 relies on abrupt-dynamics detection rather than periodic pattern recognition.
\end{abstract}

\section{Introduction}
\label{sec:intro}

Time series foundation models (TSFMs) such as Chronos-T5 \citep{ansari2024chronos}, TimesFM \citep{das2024timesfm}, MOMENT \citep{goswami2024moment}, and Moirai \citep{woo2024moirai} achieve competitive or state-of-the-art performance across diverse forecasting benchmarks, often in zero-shot settings. Yet despite rapid adoption in production systems, the internal representations of these models remain entirely unexamined from a mechanistic perspective.

In natural language processing, mechanistic interpretability (MI) has become a productive research program. Sparse autoencoders (SAEs) decompose the dense, superposed activations of language models into interpretable features \citep{anthropic2023monosemanticity,cunningham2024sparse,anthropic2024scaling}, and circuit-level analyses have identified interpretable computational subgraphs \citep{wang2022ioi,olsson2022induction,conmy2023acdc}. For time series, interpretability has instead relied on post-hoc methods: saliency maps \citep{zhao2023ts_interp_survey,kechris2025saliency}, perturbation-based explanations \citep{enguehard2023perturbations,liu2024contralsp,queen2023timex}, counterfactual approaches \citep{yan2023counterfactual}, and concept-based frameworks \citep{vansprang2024cbm_ts,boileau2025interpretable_tsfm,santamaria2025moment_latent}. Only \citet{kalnare2025mi_ts} have applied any form of mechanistic analysis to a time series transformer, targeting a small custom classifier rather than a foundation model.

To our knowledge, \textbf{no prior work has applied sparse autoencoders to a time series foundation model}. The T5 architecture underlying Chronos is well-understood \citep{raffel2020t5}, SAE training protocols are mature, and Chronos's discrete tokenization (4{,}096 bins) provides a natural unit of analysis. We address this gap with three contributions:
\begin{enumerate}[leftmargin=*,itemsep=1pt,topsep=2pt]
    \item We train TopK SAEs at six extraction points in Chronos-T5-Large and demonstrate via systematic ablation that learned features are causally relevant (392 ablations, 100\% positive CRPdeltaS ).
    \item We show that temporal concepts follow a depth-dependent hierarchy: low-level frequency features in early layers, causally critical change-detection features in the mid-encoder, and rich semantic compression in the final encoder.
    \item We find that causal importance is inversely related to semantic richness: the mid-encoder is most critical (max $\Delta\crps = 38.61$) while the final encoder paradoxically improves under progressive ablation.
\end{enumerate}

\section{Method}
\label{sec:method}

\textbf{Chronos-T5.}\quad Chronos-T5 \citep{ansari2024chronos} adapts the T5 encoder-decoder architecture \citep{raffel2020t5} for probabilistic time series forecasting via a quantization-based tokenization scheme: each value in a univariate time series is normalized and mapped to one of $V = 4{,}096$ discrete bins. Chronos-T5-Large has 710M parameters distributed across 24 encoder and 24 decoder layers with hidden dimension $d_{\text{model}} = 1{,}024$. We target six extraction points spanning the full processing pipeline: encoder blocks 5 (early), 11 (mid), and 23 (final), decoder blocks 11 and 23 (residual stream), and the cross-attention output at decoder block 11.

\textbf{Sparse Autoencoders.}\quad We train TopK sparse autoencoders \citep{cunningham2024sparse} on the residual stream activations at each extraction point. Given an activation vector $\mathbf{x} \in \mathbb{R}^{d_\text{model}}$, the SAE computes:
\begin{align}
    \mathbf{z} &= \text{TopK}\!\big(\mathbf{W}_\text{enc}(\mathbf{x} - \mathbf{b}_\text{dec}) + \mathbf{b}_\text{enc},\; k\big), \qquad
    \hat{\mathbf{x}} = \mathbf{W}_\text{dec}\,\mathbf{z} + \mathbf{b}_\text{dec}
\end{align}
where $\text{TopK}$ retains only the $k$ largest activations and zeros the rest. We set $d_\text{sae} = 8 \times d_\text{model} = 8{,}192$ features per layer, $k = 64$, and train with MSE reconstruction loss for 50{,}000 steps using Adam (learning rate $3 \times 10^{-4}$, cosine decay). Dead features are periodically resampled following \citet{anthropic2024scaling}.

\textbf{Activation Extraction.}\quad Activations are collected by registering forward hooks at each target layer during Chronos-T5 inference. We use two data sources: (i) a synthetic diagnostic suite providing ground-truth temporal properties (trends, seasonality, level shifts, frequency sweeps, heteroscedastic noise) for taxonomy validation, and (ii) the ETT benchmark \citep{zhou2021informer} for causal experiments.

\textbf{Feature Taxonomy.}\quad Each SAE feature is classified into one of eleven temporal concept categories: \texttt{trend\_up/down}, \texttt{seasonality}, \texttt{level\_shift\_up/down}, \texttt{frequency\_high/low}, \texttt{high/low\_volatility}, \texttt{noise}, and \texttt{unknown}. Classification uses Pearson correlations between each feature's activation pattern on synthetic data and the ground-truth properties of each diagnostic category. Features with maximum correlation below a threshold are assigned \texttt{unknown}.

\textbf{Causal Validation.}\quad We validate feature importance through two ablation protocols. In \textit{single-feature ablation}, we zero each feature's sparse code ($z_j \leftarrow 0$), decode to produce modified activations, patch them back into the forward pass, and measure the CRPS change \citep{gneiting2007crps}: $\Delta\crps_j = \crps_\text{ablated} - \crps_\text{original}$. In \textit{progressive ablation}, we cumulatively ablate features sorted by decoder-norm contribution ($1, 2, 4, \ldots, 64$ features) and measure CRPS at each checkpoint. Experiments use 256 context windows from ETT with prediction length 64 and 4 forecast samples. An extended configuration (1{,}024 windows, 8 samples, 200 features) was run at the final encoder for higher statistical power.

\section{Results}
\label{sec:results}

\subsection{SAE Features Are Universally Causally Relevant}

Across 392 single-feature ablation experiments spanning three encoder layers, \textbf{every ablation produced a strictly positive $\Delta\crps$} (\Cref{tab:ablation}). This establishes that each tested feature encodes information the model actively uses for forecasting, and that this information cannot be recovered from the remaining features.

The distribution of causal impact is markedly layer-dependent. At encoder block 11, the top feature (\texttt{4616}) produces a $\Delta\crps$ of 38.61 with a max-to-median ratio of 30.5$\times$, revealing an extremely heavy-tailed importance distribution where a small number of features carry disproportionate causal weight. A similarly skewed pattern appears at encoder block 5 (max/median $= 27.7\times$), while encoder block 23 shows a much more uniform distribution (max/median $= 3.9\times$ in the 64-feature sweep and $1.03\times$ in the 200-feature sweep). This power-law structure has practical implications: uniform pruning strategies would disproportionately affect the critical few features in middle layers.

\begin{table}[t]
\centering
\caption{Single-feature ablation summary across encoder layers. All features are causally relevant ($\Delta\crps > 0$). The mid-encoder (block 11) shows the highest mean and maximum impact with a heavily right-skewed distribution.}
\label{tab:ablation}
\vspace{4pt}
\small
\begin{tabular}{@{}lccccccc@{}}
\toprule
\textbf{Layer} & $n$ & \textbf{Mean} & \textbf{Med.} & \textbf{Max} & \textbf{Std} & \textbf{+Frac} & \textbf{Max/Med} \\
\midrule
Enc.\ block 5  & 64  & 3.05 & 0.95 & 26.32 & 5.12 & 1.00 & 27.7$\times$ \\
Enc.\ block 11 & 64  & \textbf{5.15} & 1.26 & \textbf{38.61} & 6.95 & 1.00 & \textbf{30.5$\times$} \\
Enc.\ block 23 & 64  & 3.73 & 2.98 & 11.65 & 2.07 & 1.00 & 3.9$\times$ \\
Enc.\ block 23$^\dagger$ & 200 & 2.37 & 2.37 & 2.44 & 0.03 & 1.00 & 1.03$\times$ \\
\bottomrule
\end{tabular}
\\[3pt]
{\footnotesize $^\dagger$Extended sweep (1{,}024 windows, 8 samples). All $\Delta\crps$ values report $\crps_\text{ablated} - \crps_\text{original}$.}
\vspace{-8pt}
\end{table}

\subsection{A Depth-Dependent Hierarchy of Temporal Concepts}

\Cref{tab:taxonomy} presents the concept distribution across all six extraction points. Of 49{,}152 total SAE features (8{,}192 per layer $\times$ 6 layers), 8{,}462 (17.2\%) receive non-unknown labels, with coverage ranging from 3.1\% at decoder block 11 to 59.8\% at encoder block 23.

\textbf{Early encoder (block 5, 4.9\% labeled).} Only 404 features receive labels, dominated by \texttt{frequency\_high} (97) and \texttt{high\_volatility} (68), suggesting preliminary local feature extraction.

\textbf{Mid-encoder (block 11, 25.8\% labeled).} This layer shows a distinctive profile dominated by \texttt{level\_shift\_up} (1{,}024 features, 12.5\%), \texttt{noise} (413, 5.0\%), and \texttt{high\_volatility} (268, 3.3\%), while seasonality is nearly absent (45, 0.5\%). The mid-encoder functions as a change-detection hub.

\textbf{Final encoder (block 23, 59.8\% labeled).} The richest layer: \texttt{seasonality} dominates (1{,}439, 17.6\%), followed by \texttt{level\_shift\_up} (1{,}097, 13.4\%), \texttt{frequency\_high} (668, 8.2\%), and \texttt{frequency\_low} (456, 5.6\%). All concept categories are represented, compressing a full temporal characterization for cross-attention.

\textbf{Decoder layers (3.1--5.5\% labeled).} All decoder points show low labeling, with \texttt{low\_volatility} and \texttt{frequency\_low} relatively prominent, possibly reflecting the decoder's focus on smooth forecast generation.

\begin{table}[t]
\centering
\caption{Feature taxonomy across layers. The final encoder (block 23) has the richest semantic coverage; the mid-encoder (block 11) concentrates change-detection features. Counts below 10 omitted.}
\label{tab:taxonomy}
\vspace{4pt}
\small
\setlength{\tabcolsep}{3pt}
\begin{tabular}{@{}lcccccc@{}}
\toprule
\textbf{Concept} & \textbf{Enc 5} & \textbf{Enc 11} & \textbf{Enc 23} & \textbf{Dec 11} & \textbf{Dec 11$_{ca}$} & \textbf{Dec 23} \\
\midrule
Seasonality       & 12  & 45   & \textbf{1{,}439} & 33 & 91  & 64 \\
Level shift $\uparrow$  & 66  & \textbf{1{,}024} & 1{,}097 & 28 & 37  & 28 \\
Level shift $\downarrow$ & 46 & --    & 210   & 14 & 26  & 60 \\
Trend $\uparrow$  & 48  & 220  & 228   & 13 & 12  & 37 \\
Trend $\downarrow$ & 11 & 16   & 89    & -- & 13  & 32 \\
Freq.\ high       & 97  & 91   & 668   & 16 & -- & 24 \\
Freq.\ low        & 10  & --   & 456   & 83 & 82  & 93 \\
High volatility   & 68  & 268  & 260   & -- & 13  & 14 \\
Low volatility    & 14  & 28   & 138   & 33 & 43  & 94 \\
Noise             & 32  & 413  & 315   & 24 & 17  & -- \\
\midrule
\textbf{Labeled (\%)} & \textbf{4.9} & \textbf{25.8} & \textbf{59.8} & \textbf{3.1} & \textbf{4.2} & \textbf{5.5} \\
\bottomrule
\end{tabular}
\\[3pt]
{\footnotesize Dec 11$_{ca}$ = cross-attention output at decoder block 11. Labels are heuristic correlation-based assignments on synthetic diagnostics.}
\vspace{-8pt}
\end{table}

\subsection{Causal Importance Is Inversely Related to Semantic Richness}

Combining ablation (\Cref{tab:ablation}) with taxonomy (\Cref{tab:taxonomy}) reveals a counterintuitive pattern. The mid-encoder, with only 25.8\% labeled features, is the most causally critical layer (mean $\Delta\crps = 5.15$, max $= 38.61$). The final encoder, with 59.8\% labeled, has lower per-feature impact and a far more uniform distribution.

\begin{figure}[t]
\centering
\includegraphics[width=0.85\linewidth]{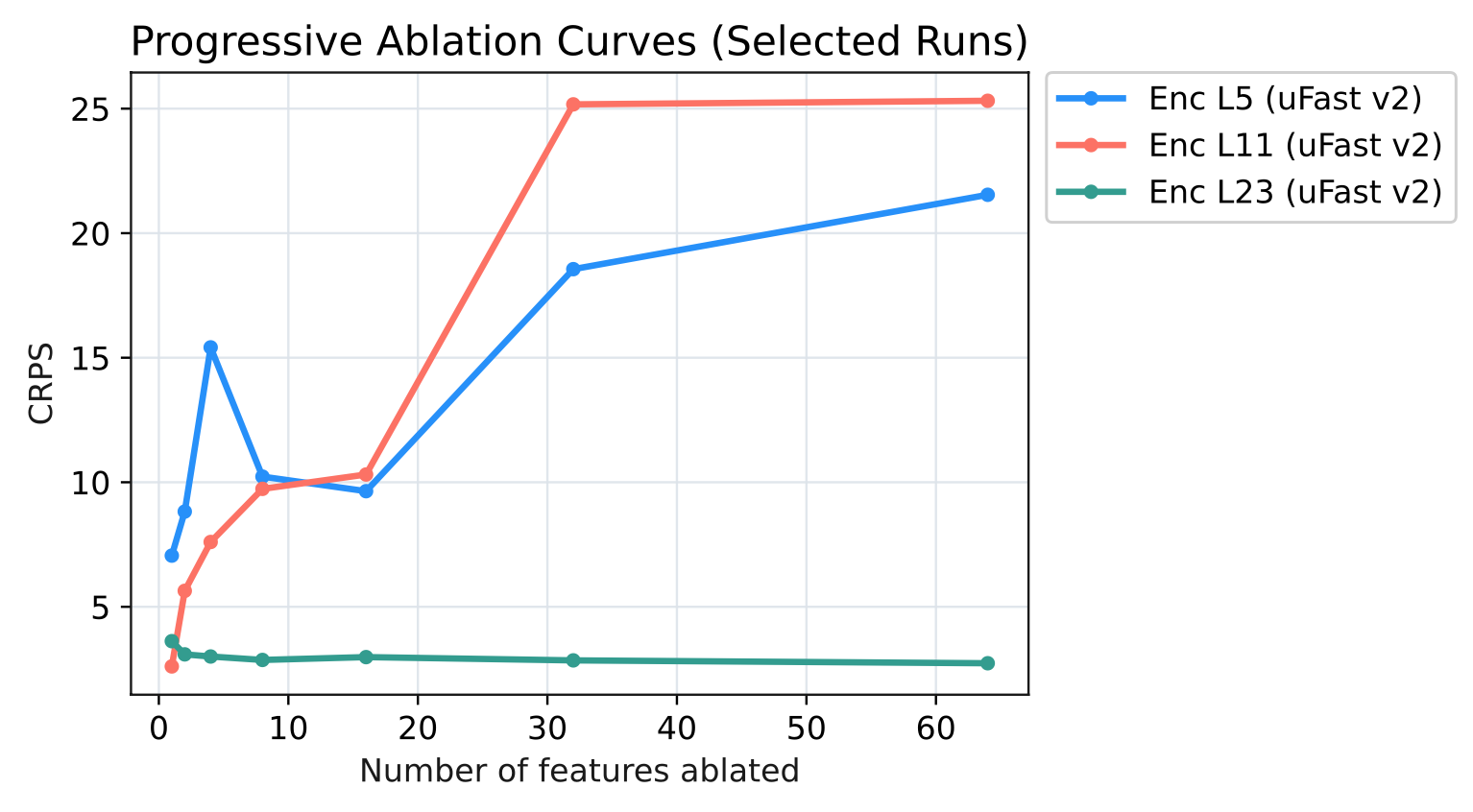}
\vspace{-8pt}
\caption{Progressive ablation curves across three encoder layers. CRPS is measured as features are cumulatively removed in order of decreasing decoder-norm contribution. Early and mid-encoder layers degrade sharply, while the final encoder remains flat or improves.}
\label{fig:overlay}
\vspace{-8pt}
\end{figure}

Progressive ablation (\Cref{fig:overlay}) makes this divergence dramatic. At encoder block 11, CRPS rises from 2.61 to 25.32 (64 features ablated), indicating catastrophic dependence. At block 5, CRPS rises from 7.05 to 21.54. At encoder block 23, however, CRPS \emph{decreases} from 3.62 to 2.73, a net improvement of 0.89. The extended 200-feature configuration confirms this: CRPS stays flat from 3.59 to 3.58 ($\Delta = -0.01$).

This paradoxical improvement may reflect the final encoder containing features that serve generalization across Chronos-T5's diverse pretraining domains but are suboptimal for ETT specifically; ablating them functions as implicit domain adaptation. The high feature utilization at block 23 (54.0\% active vs.\ 23.3\% at block 11 and 4.8\% at block 5) is consistent with representational redundancy that mildly degrades the cross-attention signal.

\section{Discussion and Conclusion}
\label{sec:discussion}

\textbf{MI transfers to time series.} Our central finding is that SAEs produce causally meaningful features when applied to a TSFM. The 100\% positive ablation rate across 392 features is consistent with the utility of SAE features as causal handles.

\textbf{The mid-encoder as computational bottleneck.} The concentration of causal importance at encoder block 11, dominated by level shifts (1{,}024 features) and noise (413), suggests that detecting abrupt distributional changes rather than periodic patterns is central to Chronos-T5's forecasting on ETT data.

\textbf{Limitations.} The taxonomy classifier is heuristic, with 82.8\% of features globally remaining unlabeled and decoder coverage below 6\%. Ablation experiments use ETT data only, and we analyze only Chronos-T5-Large. The ultra-fast ablation configuration (256 windows, 4 samples) provides directional but statistically limited findings. Future work includes SAEs at larger expansion factors, supervised probes for improved taxonomy coverage, cross-architecture comparisons, and circuit-level analyses connecting features to forecast outputs.

\bibliography{references}
\bibliographystyle{iclr2026_conference}

\appendix
\section{LLM Usage Disclosure}
\label{app:llm}

In accordance with ICLR 2026 policies on LLM usage, we disclose the following: Claude (Anthropic) and OpenAI Prism was used to assist with drafting and editing portions of this manuscript and for generating experiment infrastructure code. All experimental results, numerical findings, and scientific claims were produced by the authors' own experiments and verified independently against raw output artifacts. The authors take full responsibility for all content.

\section{Additional Experimental Details}
\label{app:details}

\textbf{SAE Training.}\quad Each SAE was trained on residual stream activations extracted from approximately 100{,}000 time series windows with a batch size of 2{,}048 and MSE reconstruction loss. TopK sparsity $k=64$ was selected for reconstruction quality while maintaining interpretable sparsity.

\textbf{Ablation Configurations.}\quad The \textit{ultra-fast} configuration (uFast v2) used 256 ETT windows, 4 forecast samples, batch size 32, features selected by decoder-norm ranking, and progressive checkpoints at $\{1, 2, 4, 8, 16, 32, 64\}$. The \textit{extended} configuration (fast v1, encoder block 23 only) used 1{,}024 windows, 8 samples, batch size 16, and checkpoints at $\{1, 2, 4, 8, 16, 32, 64, 96, 128, 160, 200\}$.

\textbf{Taxonomy Progression.}\quad The classifier was iteratively improved: Stage 0 had a wiring bug (0\% labeled), Stage 1 fixed activation loading (15.3\%), and Stage 2 (this paper) added correlation-aware directional labeling (17.2\%).

\end{document}